\documentclass{article}

\pdfoutput=1

\usepackage{graphicx} 
\usepackage{subfigure} 

\usepackage{natbib}
\usepackage{algorithm}
\usepackage{algorithmic}
\usepackage{amsmath,amssymb,graphicx}
\usepackage{bbm}

\usepackage[draft]{hyperref}

 \usepackage[accepted]{icml2013}

\icmltitlerunning{Dependent partition-valued processes}

\begin{document} 

\twocolumn[
\icmltitle{A dependent partition-valued process for multitask clustering and time evolving network modelling}

\icmlauthor{Konstantina Palla \textdagger}{kp376@cam.ac.uk}
\icmladdress{University of Cambridge}
\icmlauthor{David A. Knowles \textdagger}{davidknowles@cs.stanford.edu}
\icmladdress{University of Cambridge}
\icmlauthor{Zoubin Ghahramani}{zoubin@eng.cam.ac.uk}
\icmladdress{University of Cambridge}

\icmlkeywords{clustering, Bayesian nonparametrics}

\vskip 0.3in
]

\begin{abstract} 

The fundamental aim of clustering algorithms is to partition data points. We consider tasks where the discovered partition is allowed to vary with some covariate such as space or time. One approach would be to use fragmentation-coagulation processes, but these, being Markov processes, are restricted to linear or tree structured covariate spaces. We define a partition-valued process on an arbitrary covariate space using Gaussian processes. We use the process to construct a multitask clustering model which partitions datapoints in a similar way across multiple data sources, and a time series model of network data which allows cluster assignments to vary over time. We describe sampling algorithms for inference and apply our method to defining cancer subtypes based on different types of cellular characteristics, finding regulatory modules from gene expression data from multiple human populations, and discovering time varying community structure in a social network. 

\end{abstract} 

\section{Introduction}
\label{sec:introduction}
We are interested in problems where the partitioning of data into groups depends on some covariate. As a simple example, consider how the partitioning of a number of people into friendship groups might evolve over time. At proximal times people will tend to form similar partitions, but at more distant times the partitioning might be quite different. We define a nonparametric process that induces dependency between partitions on an arbitrary covariate space. 

Many nonparametric processes studied in the literature, such as the Dirichlet process~\citep[DP,][]{Ferguson1973}, are distributions over the space of measures.  The DP can be used to construct a distribution over the space of partitions known as the Chinese restaurant process~\citep[CRP,][]{Aldous1983}. Dependent nonparametric processes extend distributions over measures and partitions to give distributions over collections of measures or partitions indexed by locations in some covariate space~\cite{maceachern1999dependent}. Covariate spaces include $\mathbb{R}^+$(e.g. continuous time), $\mathbb{Z}$ (e.g. discrete time), or $\mathbb{R}^d$ (e.g. geographical location). Most dependent nonparametric processes define distributions over collections of measures. Dependency among the measures at each location may be induced in various ways. The single-p Dependent Dirichlet Process~\citep[DDP,][]{maceachern99} assumes a shared set of atom weights but allows the atom locations to vary e.g. according to a Gaussian process~\citep[GP,][]{rasmussen06}. The multiple-p DDP \citep{maceachern00} extends this construction by allowing the atom weights to vary across the covariate space. 

By de Finetti's theorem, any exchangeable sequence is equivalent to i.i.d. draws from a conditional distribution and can be written as a mixture of such distributions. Both the single-p and multiple-p DDP have the DP as their de Finetti mixing measure and introduce dependency in the mixing distribution either through the atom locations or weights. The generalized spatial Dirichlet process (GSDP) by \citet{Duan05generalizedspatial} induces dependency by assuming that the mixing distribution is common to all covariate indices, but the conditional distributions, that is the distribution of the assignments at each covariate given the draw from the DP, are correlated. 
Marginally at each covariate location, the process is a DP mixture of Gaussians. 
A similar method is used in the dependent Pitman-Yor process of~\citet{sudderth08a} to segment images. In both models, the authors link assignments via thresholded GPs. In this paper, we propose a process that correlates the partitioning among the covariates using the same idea of thresholded GPs. However, while Duan et al. and Sudderth and Jordan focus on modelling spatial random effects and spatial image segmentation tasks respectively, we generalise these ideas and apply them in multitask clustering and time evolving network modelling. 

Despite this long line of research into dependent measure-valued processes, little attention has been given to their interpretation as dependent \emph{partition}-valued processes. A sample from such a process is collection of partitions indexed by the covariate: at any single covariate location, there is a single partition. An exception is \citet{TehBluEll2011a}, where the duality between Kingman's coalescent 
 and the Dirichlet diffusion tree 
 is leveraged to define a ``Fragmentation-Coagulation'' process (FCP) which is Markov, stationary, exchangeable and has CRP distributed marginals \citep{BertoinFcp}. The FCP defines a distribution over a collection of partitions on a one dimensional covariate space. The partitioning at each covariate location is a result of fragmentation (according to the DDT) and coagulation (according to Kingman's coalescent) events that take place between adjacent covariate locations. Although mathematically elegant, due to its Markov construction it is not clear how to extend the FCP to an arbitrary covariate space. In this paper, we derive a dependent partition-valued process on an arbitrary covariate space which, like the FCP, is exchangeable and has CRP distributed marginals. For brevity, we refer to this process as DPVP for ``Dependent Partition-Valued Process''. 

The DPVP is also closely related to the dependent IBP \citep[dIBP,][]{williamson10}, which addresses the problem of modelling dependence for binary latent feature models. Coupling over the covariate, in both processes, is achieved by representing Bernoulli variables at each covariate index as transformed Gaussian variables and aggregating these into Gaussian processes over the covariate. The dIBP generates a set of binary feature matrices evolving over the covariate, while the DPVP couples a set of random partitions.   

We use the DPVP to construct two distinct models. The first is a multitask clustering model (MCM) which attempts to find similar partitions of objects across distinct data views. Our approach learns the similarity between the clustering in each data source. MCM is closely related to the Multiple Dataset Integration model~\citep[MDI,][]{KirkGSGW12}, where the conditional probability of allocating a sample to a cluster in one data source is influenced by the assignments in other data sources. Like MCM, MDI can learn how similar the clusterings should be across different data sources using positive real-valued parameters for each pair of data sources. However, MCM is a valid generative model unlike MDI which is defined only in terms of these conditional distributions. Both models can be considered as using finite approximations to a DP mixture model at each location (data source): where MDI simply uses a large finite Dirichlet distribution, MCM uses the stick breaking construction of the DP~\cite{
sethuraman94}. 

The second model we construct is an evolving community structure model (ECS) for time series network data. The latent partitioning at each time point determines the links among the objects and is a time series extension of the Infinite Relationship Model~\citep[IRM,][]{Kemp2006}. The literature on modelling dynamic network structure is large but only few models follow a Bayesian approach~\citep{Xing10,heaukulani2013dynamic}. \citet{Xing10} for example defines a dynamic mixed membership stochastic block (dMMSB) model, where the prior weights and parameters are given by softmax transformations of linear Gaussian state space models. Each person is associated with a mixed membership vector which is conditionally independent at each time point given the time evolving prior weights over the roles. In contract, in ECS each person belongs to only one cluster at each time point, with dependency introduced directly between these assignments through time. Thus ECS, unlike dMMSB, can encode, for example, that an individual in a particular friendship group at time $t$ is more likely to remain in that group at time $t+1$. 

In Section 2, we briefly provide some background on partitions, the CRP and the stick breaking construction. In Section 3, we present the dependent partition-valued process and in Sections 4 and 5 describe how to use this process to build a multitask clustering model and a model for time evolving community structure respectively. In Section 6, we describe how different choices of kernel may be used for different applications. Inference in our model is performed via a Gibbs sampler, which is described in Section 7. In Sections 8 and 9 we describe case studies for multitask clustering and network modelling. Finally, in Section 10 we conclude our work and discuss some future directions.

\section{Background}

A partition of $[N]=\{1,..,N\}$ is a set of disjoint non-empty subsets of $[N]$ such that the union of these subsets is $[N]$. In clustering applications we refer to each subset as a ``cluster''. The set of partitions of $[N]$ is denoted $\Pi_N$. The most natural distribution over $\Pi_N$ is the Chinese restaurant process (CRP). Since the CRP is exchangeable (invariant to permutations of $[N]$), by de Finetti's theorem it can be represented using i.i.d. samples from a random measure. For the CRP this is the Dirichlet process (DP). The weights of the DP can be represented using a ``stick breaking'' construction as follows~\cite{sethuraman94, Ishwaran01}:
\begin{align} \label{eq:stickbreaking}
\pi_k & = v_k \prod_{l=1}^{k-1} (1-v_l)  &\forall k \in \mathbb{Z} \nonumber \\
v_k &\sim_\text{iid} \text{Beta}(1,\alpha) &\forall k \in \mathbb{Z}   
\end{align}
In the above, the $\pi_k$'s are mixture proportions (stick lengths) and $\alpha>0$ is known as the concentration parameter. If we now consider sampling cluster assignments $c_n|\pi \sim \text{Multinomial}(\pi)$ for $n \in [N]$, then the corresponding partition $\gamma$ of $[N]$ is marginally CRP distributed. The partition is obtained from the cluster assignments $c$ by putting $n$ and $m$ in the same subset iff $c_n=c_m$. 

\section{A dependent partition-valued process} 

Given some covariate space $\mathcal{T}$, our aim is to construct an exchangeable, projective, dependent process $(\gamma(t),t \in \mathcal{T})$ such that each $\gamma(t)$ is a random partition which is marginally CRP distributed. To achieve this we generalize the process proposed by  \citet{Duan05generalizedspatial} and \citet{sudderth08a} as follows. 

We will use the following representation of the stick-breaking construction (Equation~\ref{eq:stickbreaking}):
\begin{enumerate}
 \itemsep-0.2em
 \item[] For each object $n \in [N]$
 \item Set $k:=1$
 \item Sample $a_{nk} \sim \text{Bernoulli}(v_k)$
 \item If $a_{nk}$ then assign $c_n:=k$, else increment $k$ and go to 2. 
\end{enumerate}
It is straightforward to see that the probability of choosing cluster $k$ is given by the stick length $\pi_k$ since 
\begin{align}
P(c_n= k|v) &= P(a_{n1}=0|v)\dots \nonumber \\ 
	& P(a_{n{(k-1)}}=0|v)P(a_{nk}=1|v) \nonumber \\
&= (1-v_1)\dots (1-v_{k-1}) v_k = \pi_k
\end{align}

We note that the Bernoulli random variable $a_{nk}$ can be represented as
\begin{align}\label{eq:invbern}
f_{nk} &\sim \mathcal{N}(0,\sigma^2)\nonumber \\
 a_{nk} &= \mathbb{I}[ f_{nk} < \phi^{-1}(v_k|0,\sigma^2) ]
\end{align}
where $\phi(.|\mu,\sigma^2)$ is the Gaussian cumulative distribution function with mean $\mu$ and variance $\sigma^2$. We now extend these random variables to random functions on $\mathcal{T}$ and introduce dependency by extending the Gaussian prior on each $f_{nk}$ to a Gaussian process (GP) prior on each $f_{nk}(t)$ with $t\in \mathcal{T}$:
\begin{align}\label{eq:gpgen}
 f_{nk}(t) &\sim_\text{iid} \text{GP}(0,\Sigma(t,t')) & \forall n \in [N], k \in \mathbb{Z} \nonumber \\
 a_{nk}(t) &= \mathbb{I}[ f_{nk}(t) < \phi^{-1}(v_k|0,\Sigma(t,t)) ] 
\end{align}
where $\Sigma(t,t')$ is the covariance function which we assume to be common to all the GPs. As a result of the marginalisation properties of GPs, each $f_{nk}(t)$ is marginally $N(0,\Sigma(t,t))$ distributed, so that $a_{nk}(t)$ is marginally Bernoulli$(v_k)$ distributed, and the resulting partition $\gamma(t)$ is marginally CRP distributed, as desired. 

To summarise, the DPVP generative process is
\begin{align}\label{eq:gp}
v_k &\sim_\text{iid} \text{Beta}(1,\alpha) &\forall k \in \mathbb{Z} \nonumber \\
f_{nk} &\sim_\text{iid} \text{GP}(0,\Sigma) & \forall n \in [N], k \in \mathbb{Z} \nonumber \\
c_n(t) &= \min_{ f_{nk}(t) < \phi^{-1}(v_k|0,\Sigma(t,t)) } k 
\end{align}
We do not make the $v_k$ a function on $\mathcal{T}$ but instead assume global mixing proportions: relaxing this constraint would be a straightforward extension. Moreover, by setting $v_k \sim_\text{iid} \text{Beta}(1-d,\alpha + kd)$, where $\alpha>-d \text{ and } 0 \leq d <1$, we could easily obtain Pitman-Yor marginals.

In practice, we cannot represent a countably infinite set of GPs. While we could adaptively extend our representation as required, we instead choose the simpler option of truncating the stick breaking construction at some level $K$. If $a_{nk}(t)=0$ for all $k < K$, then we set $c_n(t)=K$. We use vector notation $\mathbf{v}=\{v_1, \dots, v_{K-1}\}$ for the set of the beta distributed stick length parameters in Equation \ref{eq:gp}. To sample from the $K$-truncated DPVP at $T$ locations $\{t_\tau \in \mathcal{T} | \tau=1,..,T\}$ we require $K-1$ $T$-vectors $\mathbf{f}_{nk} = [f_{nk}(t_1) \dots f_{nk}(t_T)]$, for each object $n$, from a Gaussian process with a $T \times T$ Gram (covariance) matrix, $\mathbf{\Sigma}$, where $\Sigma_{\tau \tau'} = \Sigma(t_\tau,t_{\tau'})$. 
For notational simplicity, we concatenate the $f$ vectors into a $T \times N(K-1)$ matrix, $\mathbf{F}$. By drawing from a GP as in Equation \ref{eq:invbern}, we introduce dependency among the partitions in different covariate locations. The dependence is defined by the covariance function of the GP, $\mathbf{\Sigma}$, and introduces similarity between the partitions at proximal covariate locations. 

We denote a sample from the DPVP as $DPVP(\alpha,\mathbf{t},\Sigma)$, where $\alpha$ is the concentration parameter of the underlying DP, $\mathbf{t}$ is the vector of covariate locations and $\Sigma$ is the Gram matrix. 
Having presented the process, we can describe how it relates to~\citet{sudderth08a}. 
The non-hierarchical version of Sudderth and Jordan's model can be derived by setting $N=1$ in the proposed model. 
Although our model is inspired by the same construction, their model does not partition data points since $N=1$. 

\begin{figure} 
 \includegraphics[width=0.9\columnwidth, trim = 0 0 0 0 ]{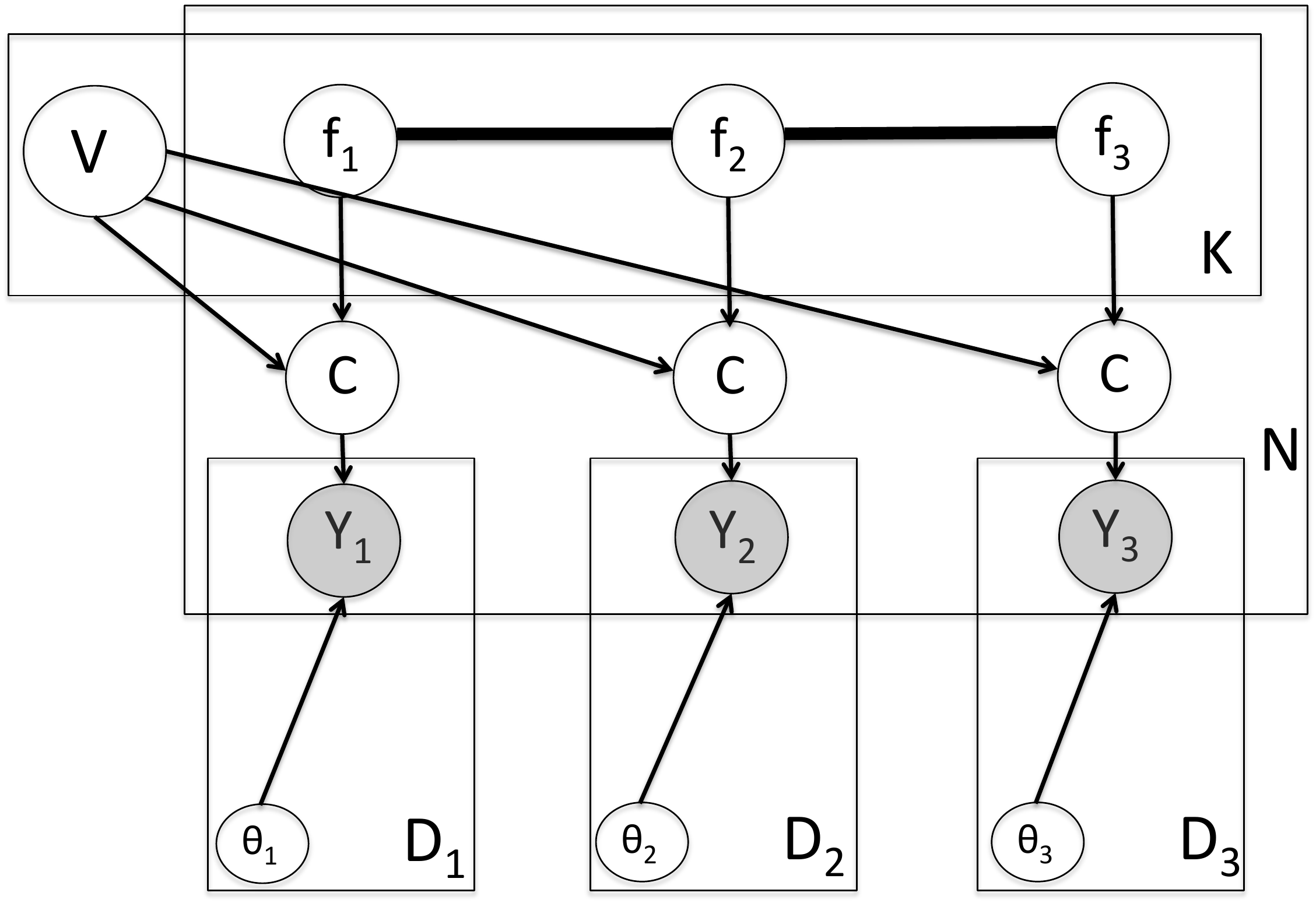}
\caption{Graphical model for both the Multitask Clustering Model (MCM) and Evolving Community Structure (ECS) model. $c$ are a deterministic function of $f$ and $v$ but we show both for clarity. The thick line linking the $f$'s denotes dependence through the Gaussian process.}\label{fig:genM}
\end{figure}

We now use this process to construct two models: a multitask clustering model and a network model that allows the discovered community structure to vary through time. The graphical model for both is shown in Figure~\ref{fig:genM}.

\section{Multitask clustering} 

We are interested in the situation where we have a collection of $N$ objects, for each of which we have measurements from $T$ different data sources. We associate each data task $\tau$ with a covariate location $t_\tau \in \mathcal{T}$, although this might be latent. 
Our model will assume that for each data source the objects are grouped into clusters forming a partition. We allow the clustering to be different for each data source, but model dependency between these partitions using the DPVP. Denote the data for object $n$ in data source $\tau$ as $y_n^\tau \in \mathcal{Y}^\tau$, where we allow the observed space $\mathcal{Y}^\tau$ to be different for each data source. The Multitask Clustering Model (MCM) is then 
\begin{align}\label{eq:mcm}
c_n^\tau &\sim DPVP(\alpha, \mathbf{t}, \Sigma) \nonumber \\
\theta_k^\tau &\sim G^\tau \nonumber \\
y_n^\tau | c_n^\tau, \theta_k^\tau &\sim F^\tau(\theta^\tau_{c_n^\tau})
\end{align}
where $\theta_k^\tau$ are cluster parameters, $G^\tau$ are priors on the cluster parameters and $F^\tau$ are data likelihoods. In the following we assume all the data sources are continuous, i.e. $\mathcal{Y}^\tau = \mathbb{R}^{D^\tau}$, and can therefore be represented as a $N \times D^\tau$ matrix $\mathbf{Y}^\tau \in \mathbb{R}^{N \times D^\tau}$. We allow each data source to have a different observed dimensional $D^\tau$. As a simple concrete example of this model, we use a diagonal (independent per dimension) Gaussian likelihood and its conjugate normal-gamma prior on the cluster parameters:
\begin{align} 
(\mu^\tau_{k d} ,\lambda^\tau_{k d}) &\sim \mathcal{N}(\mu^\tau_{k d}| \mu_o, \frac{1}{\kappa_o\lambda^\tau_{k d}}) \mathcal{G}a(\lambda^\tau_{k d} | \alpha_o, \frac{1}{\beta_o})\nonumber \\ 
y_{nd}^\tau|c_n^\tau,\mu ,\lambda &\sim \mathcal{N}(\mu^\tau_{c_n^\tau d},1/\lambda^\tau_{c_n^\tau d}) \label{eq:gausslike}
\end{align}
where we set $\kappa_0=0.1,\mu_0=0,\beta_0=0.1,\alpha_0=0.1$. The cluster paramaters $\theta_k^\tau$ in Equaion \ref{eq:mcm} are now $\theta_k^\tau = (\mu^\tau_{k} ,\lambda^\tau_{k})$. By choosing the conjugate prior we are able to integrate out the cluster parameters during inference. The choice of a diagonal covariance allows scaling to larger datasets. The generalisation to other data types would be straightforward using the appropriate conjugate prior.

\section{A model for time evolving community structure}

Most models of network data assume a static network, whereas real world networks typically evolve over time. We use the DPVP to build a model which discovers clusters in network data, but allows cluster assignments to change over time. We refer to this model as ECS for ``Evolving Community Structure''. In this case our data $\mathbf{Y}^\tau$ at each location $\tau$ is a binary $N\times N$ matrix representing the presence or absence of links between objects. The assignment of the objects to groups at each covariate index $t$ determines the probability of links in an analogous fashion to the Infinite Relationship Model~\citep[IRM,][]{Kemp2006}. 
\begin{align}
c_n^\tau &\sim DPVP(\alpha, \mathbf{t}, \Sigma) \\
\theta_{kk'}^\tau &\sim \text{Beta}(\beta,\beta) \\
y_{nn'}^\tau | c^\tau, \theta^\tau &\sim \text{Bernoulli}(\theta^\tau_{c_n^\tau c_{n'}^\tau})
\end{align}
where $y_{nn'}^\tau$ denotes the presence of a link between objects $n$ and $n'$ at time $\tau$, and we set $\beta=0.1$ to encourage values of $\theta$ close to $0$ or $1$. While we assume the link probabilities $\theta$ are independent at every time point $\tau$ a possible extension would be to introduce dependence between these values at adjacent time points. 

\section{Choice of kernel} \label{sec:kernels}

Since the DPVP is constructed using Gaussian processes there is great flexibility in the choice of covariance function (kernel). Here we describe only the options used in the experiments in Section~\ref{sec:experiments}. We choose to fix the diagonal $\Sigma(\tau,\tau)=1$ in all cases since the DPVP is invariant to scaling of the covariance matrix: by construction changing the prior marginal variance of the GP functions at any location does not effect the resulting distribution over partitions. 

\paragraph{Squared exponential.} In the situation where there is a known covariate value, such as time or spatial location, associated with each data source or network, the canonical covariance function is the squared exponential kernel
\begin{align*}
\Sigma(t, t') = \exp\left(- \frac{(t-t')^2}{2l^2}\right)
\end{align*}
where $l>0$ is the lengthscale which controls the smoothness of the GPs. We put an exponential prior on $l$. 

\paragraph{Similarity kernel.} In the multitask clustering setting we may often have little prior knowledge about which data sources are likely to have similar clustering structure to others and would like to learn this from the data itself. In this case we fix the diagonal terms of the kernel $\Sigma$ equal to $1$ and put a Uniform$[-1,1]$ prior on the off-diagonal terms. We ensure PSD matrices simply by rejecting any non-PSD matrices during the slice sampling. 

\paragraph{Tree structured covariance.} In other cases we may have a tree structured dependency between the data sources: the data sources are the leaves of a tree which represents a known relationship. Each branch represents a Gaussian factor: the child node is normally distributed with mean equal to its parent and variance equal to the branch length. The total height of the tree is $1$ so that the diagonal of the resulting covariance matrix is $1$. We put a uniform prior on the branch lengths (under the constraint that no branch lengths can be negative). A concrete example is given in Section~\ref{sec:hapmap}. 

\section{MCMC Inference}
\label{inference}
We present a method for inferring the latent variables and parameters of MCM and ECS models: the matrix of Gaussian process function values, $\mathbf{F}$, the stick length parameters $\mathbf{v}$ and the parameters of the covariance matrix $\Sigma$. Exact inference is intractable, so we develop a Markov Chain Monte Carlo (MCMC) procedure to sample from the posterior distribution. Each iteration is $O(DNKT+T^3)$ which for small $T$ is the same as EM or sampling for a DPM. The cluster assignments $c$ are not represented since these are a deterministic function of $\mathbf{F}$ and $\mathbf{v}$ (since we ensure that $\Sigma_{\tau \tau'}=1$ in all cases the assignments do not depend on $\Sigma$). For both models we are able to integrate out the cluster parameters, $\theta$ (the link probabilities for ECS), due to conjugacy: 
\begin{align}
p(\mathbf{Y}^\tau|\mathbf{F}^\tau, \mathbf{v})= \int  p(\mathbf{Y}^\tau|\mathbf{F}^\tau, \mathbf{v}, \boldsymbol\theta^\tau)p(\boldsymbol\theta)d\boldsymbol\theta
\end{align}
The sampler iterates as follows:

\paragraph{Sampling the GP function values, $\mathbf{F}$.} The conditional posterior over the matrix $\mathbf{F}$ is given by
\begin{align*}\label{eq:postF}
p(\mathbf{F}| \mathbf{Y}, \mathbf{v}, \Sigma) \propto \left( \prod_{\tau=1}^T p(\mathbf{Y}^\tau|\mathbf{F}^\tau, \mathbf{v}) \right) \prod_{n=1}^N \prod_{k=1}^{K-1} N(\mathbf{f}_{nk}|0,\Sigma)
\end{align*}
Since we have a GP prior over $\mathbf{F}$ we use elliptical slice sampling \citep[ESS,][]{murray2010}, which is specifically designed to sample from posteriors with strongly correlated Gaussian priors. We find that jointly sampling the $K-1$ GPs associated with each datapoint in turn gives better mixing than attempting to jointly sample all $N(K-1)$ GPs at once (see Section~\ref{sec:mcm-syn}). The covariance matrix for the $K-1$ GPs for a single datapoint is block diagonal where each block is $\Sigma$. Naive computation of the Cholesky would be $O(K^3T^3)$, but utilising the block diagonal structure it is only $O(T^3)$, and $T$ is typically relatively small in the applications we envisage.  


\paragraph{Sampling the weight vector, $\mathbf{v}$.} 
The sampler successively samples each of the $K-1$ weights $v_k$. Since we do not have conjugacy (due to the complex form of the likelihood function), we cannot sample directly from the posterior $v_k | \mathbf{Y}, \mathbf{F}, \mathbf{v}_{-k}$ (where $\mathbf{v}_{-k}$ denotes the values of $\mathbf{v}$ excluding $v_k$). To overcome this problem, we use slice sampling~\citep{Neal2003} with the reparameterisation $g(v)= \log[v/(1-v)]$ so that $g \in \mathbb{R}$. 

\paragraph{Sampling parameters of the kernel.} Learning the parameters of the kernel is of interest because it tells us how similar the partitions appear to be across covariate space. Since we ensure $\Sigma_{\tau \tau'}=1$ the likelihood for $\Sigma$ is just $ \prod_{n=1}^N \prod_{k=1}^{K-1} N(\mathbf{f}_{nk}|0,\Sigma)$. In fact, since values of $\mathbf{f}_{nk}$ for $k > \max c$ do not effect the partitioning, we can trivially integrate these out so the product over $k$ in the likelihood need only go up to $\max c$. This both saves computation and improves mixing by avoiding conditioning on irrelevant information. For all three kernels described in Section~\ref{sec:kernels} we use slice sampling to learn the kernel parameters: the lengthscale for the squared exponential kernel, the correlation coefficients for the similarity kernel, and the branch lengths for the tree structured kernel. 


\paragraph{Sampling the concentration parameter, $\alpha$.} We also use slice sampling to infer the hyperparameter $\alpha$ using a Gamma prior $\alpha  \sim \mathcal{G}(1,1)$. The likelihood is $\prod_{k=1}^{K-1} \text{Beta}(v_k|1,\alpha)$. 

\paragraph{Initialisation.} The DPVP can suffer from the label switching problem: although the partitioning at two locations might be similar, they may look quite different according to the model if the labels are permuted. To alleviate this problem we initialised both MCM and ECS using the equivalent DPM model where the same clustering is shared across all locations. 

\section{Multitask clustering results} \label{sec:experiments}

Using synthetic data we first demonstrate some encouraging characteristics of our inference method. We then apply MCM to two real world biological datasets. Unless otherwise stated we use $1000$ MCMC iterations, discarding the first $500$ as burnin.

\subsection{Synthetic data} \label{sec:mcm-syn}

\begin{figure}[h]
\centering
\subfigure[]{ \includegraphics[width=0.8\columnwidth]{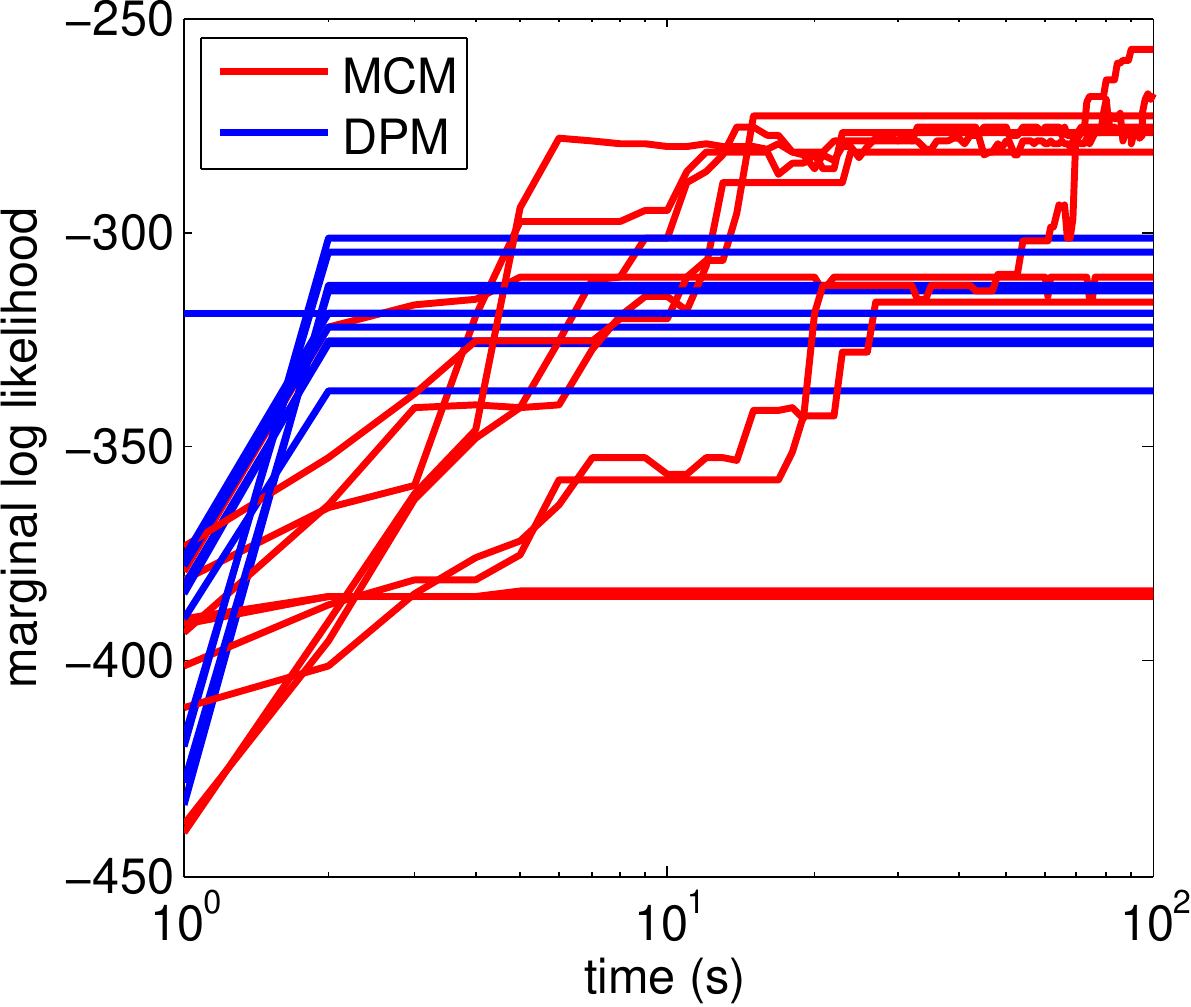}}
\subfigure[]{\includegraphics[width=0.8\columnwidth]{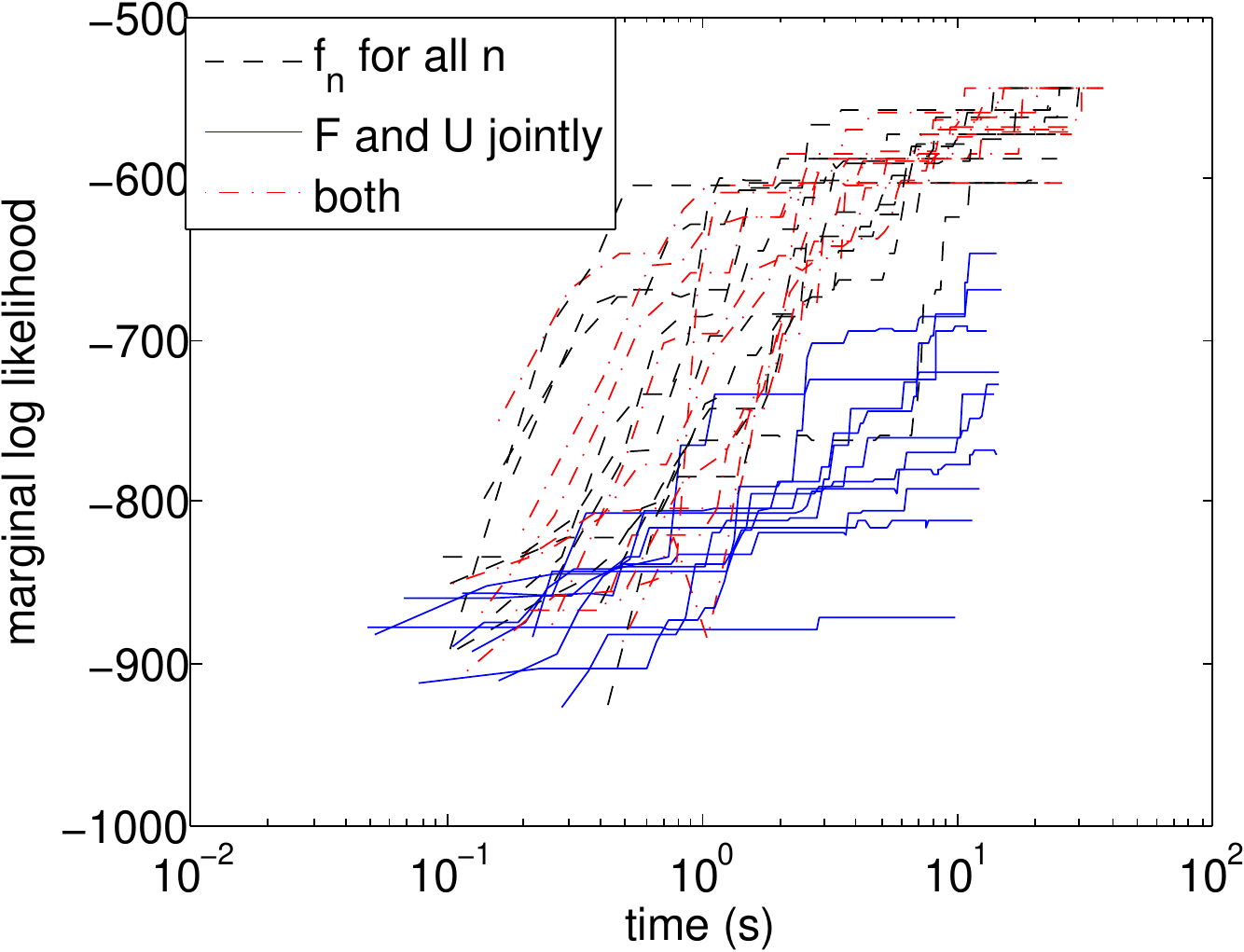}}
\vspace{-0.1in}
\caption{(a). Comparison of our ESS based inference vs standard Gibbs sampling for the DPM when there is only $T=1$ location (b). {Comparison of sampling methods for $\mathbf{F}$ across 10 MCMC runs each. Dashed black: sequentially sampling the $K-1$ GPs associated with each object $n$ sequential, followed by $v$. Solid blue: sampling all of $F$ and $U$ jointly. Dashed red: alternating between these two.}
\label{fig:mcm_comp}}
\end{figure}

We use experiments on synthetic data to demonstrate three things. Firstly, on a single location dataset with $N=30$ objects in three equal sized, well separated clusters with means $-3\boldsymbol 1, \mathbf{0}, +3 \boldsymbol 1$ and equal covariances $I$ in $D=5$ dimensions, we show that our Elliptical Slice Sampling (ESS) based inference is competitive with standard Gibbs sampling for DPMs (for $T=1$ MCM is exactly equivalent to a DPM). In Figure~\ref{fig:mcm_comp}(a) we see that while ESS does sometimes get stuck in a local mode (two of ten repeats), it typically finds areas of higher marginal likelihood than the standard Gibbs sampler. 

Secondly, on a dataset with $T=3$ data sources, two of which have the same clustering structure as the first example, and one of which has a single cluster with mean $\mathbf{0}$ and variance $I$, we show that sequentially jointly sampling the $K-1$ GPs associated with each datapoint mixes better than attempting to sampling all $N(K-1)$ GPs and $\mathbf{v}$ jointly (Figure~\ref{fig:mcm_comp}(b)). The latter approach attempts to make very large global moves in the MCMC state space so many likelihood evaluations are required before a new point is accepted by ESS. 

Thirdly, on the same dataset we show we can learn a sensible similarity kernel. We use the similarity kernel defined in Section~\ref{sec:kernels}. We run MCMC for $200$ iterations and, finding the chain appears to have burnt in after $100$ iterations, calculate 95\% credible intervals for $\Sigma$ using the final $100$ samples. The correlation coefficient between the two data sources whose true clusterings are identical has credible interval $[0.31,0.65]$, whereas between the distinct data source and these two, the credible intervals are $[-0.39,0.27]$ and $[-0.37,0.23]$, covering zeros. 

\subsection{Cancer cell line encyclopedia}

The Cancer cell line encyclopedia~\citep[CCLE,][]{barretina2012cancer} is a recently published resource to aid the understanding of why certain cancer subtypes are resistant to particular drugs. $N=432$ cancer cell lines were grown in the presence of $D_\text{drugs}=24$ different therapeutic compounds at nine different concentrations, and their growth measured after 10 days. These growth curves are summarised in terms of ``active area'': the total inhibition of growth summed over all concentrations. Alongside these sensitivity measurements various molecular characteristics of the cell lines are measured, including gene expression (GE), copy number variation (CNV, the number of times a gene is duplicated in the cancer genome) and oncogene mutations (mutations such as insertions, deletions or single nucleotide polymorphisms, SNPs, in $D_\text{onco}=1600$ genes known to be involved in cancer). For GE and CNV we just take the $D_\text{GE}=D_\text{CNV}=1000$ high variance genes. 

We use MCM to perform multitask clustering of the $N$ cell lines across the four data matrices: GE ($N \times D_\text{GE}$), CNV ($N \times D_\text{CNV}$), oncogene mutations (ONCO, $N \times D_\text{ONCO}$) and sensitivity (SENS, $N \times D_\text{drugs}$). For simplicity we use the diagonal covariance Gaussian likelihood of Equation~\ref{eq:gausslike} for each data source, which each dimension preprocessed to have zero mean and unit variance (we leave the use of different likelihoods for this heterogenous data to future work). 

We are interested in two aspects of the analysis: firstly, how well can we predict drug sensitivity, since this is clinically relevant (being able to predict sensitivity could help determine what drug is most appropriate for a particular patient) and secondly, how similar is the clustering across the four data sources. 

\begin{figure}[h]
\centering
\includegraphics[width=.8\columnwidth]{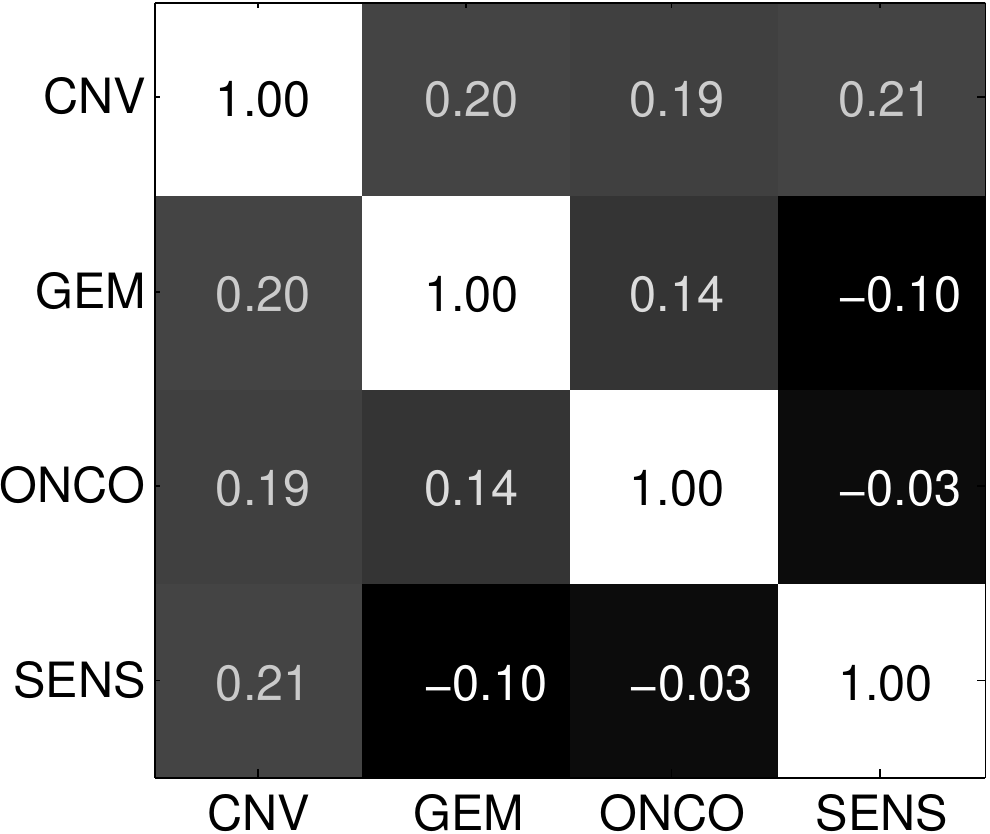}
\caption{Similarity kernel learnt for different data sources in the CCLE dataset.} \label{fig:ccle_sigma}
\end{figure}

To assess MCM at predicting drug sensitivity from molecular characteristics (GE, CNV and ONCO), we randomly choose 20 different sets of 10\% of the drug sensitivity measurements to hold out, and attempt to impute these values. We compare to two extremes: using a DP mixture (DPM) model independently on each data source (in this case we need only run the algorithm for the sensitivity data source since this is what we are interesting in imputing), and a DPM where the clustering is common to all data sources (all three methods use the same likelihood). These are extremes of MCM, representing minimum or maximum transfer learning respectively. The results are shown in Table~\ref{tab:results}. We see that the independent clustering performs very poorly, the shared clustering does reasonably well, but MCM performs best since it is able to learn how much information to transfer from the other data sources to the drug sensitivity clustering task. 
The learnt correlation matrix is shown in Figure~\ref{fig:ccle_sigma}: of particular interest are the correlation coefficients between sensitivity and the other data sources. We see there is a positive correlation to CNV, whereas the correlation is small (and in fact slightly negative) to gene expression and oncogene mutations, suggesting that copy number variation is the most indicative characteristic of which drugs a cancer will be sensitive/resistant to. 

\begin{table}
\caption{Predictive performance results for both MCM and ECS on real world datasets. Values are log predictive likelihood per heldout data entry. vdB indicates the van de Bunt's dataset.} 
\label{tab:results}
\begin{center}
\begin{small}
\begin{sc}
\begin{tabular}{p{0.98cm}p{1.99cm}p{1.99cm}p{1.99cm}}
\hline
\abovespace\belowspace
Dataset & Independent & Shared & MCM/ECS \\
\hline
\abovespace
CCLE &	$-3.221 \pm 0.552$ & $-1.109 \pm 0.069$ & $\mathbf{-0.902 \pm 0.097}$\\
vdB & $ -0.530 \pm 0.025 $ & $ -0.502 \pm 0.022 $ & $ \mathbf{-0.095 \pm 0.017} $ \\
\belowspace
HapMap & $ -1.357 \pm 0.055 $ & $ \mathbf{-1.134 \pm 0.013} $ & $ -1.277 \pm 0.016 $ \\
\hline
\end{tabular}
\end{sc}
\end{small}
\end{center}
\vskip -0.1in
\end{table}

\subsection{HapMap gene expression data} \label{sec:hapmap}

\begin{figure}[h]
 \includegraphics[width=.9\columnwidth]{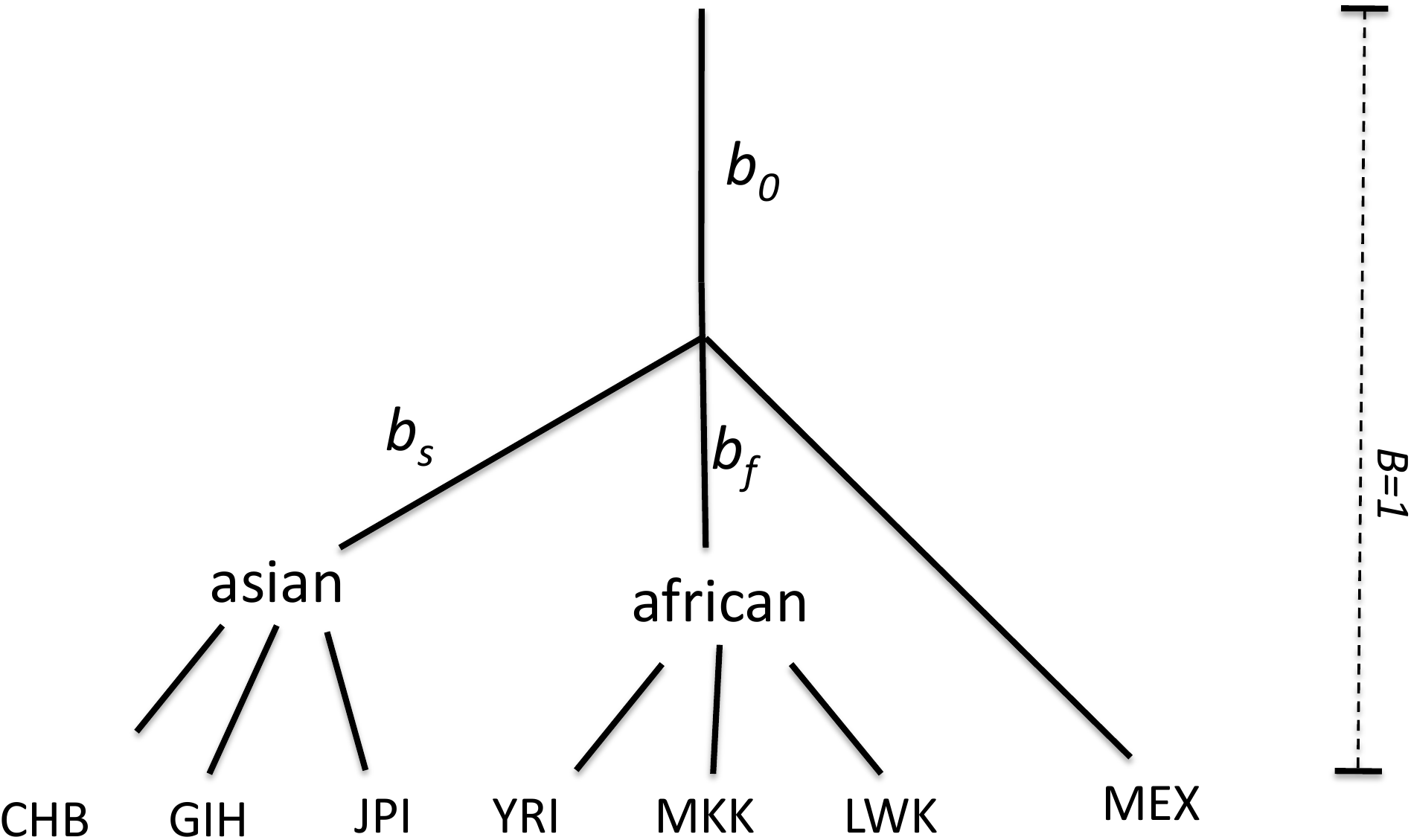}
\caption{Tree structure of human populations in HapMap. }\label{fig:popTree}
\end{figure}

The HapMap project\footnote{\url{http://www.sanger.ac.uk/resources/downloads/human/hapmap3.html}} is primarily an attempt to measure genetic variation between $1301$ individuals from different human populations, but gene expression data is also available in $D=618$ individuals~\cite{montgomery2010transcriptome}. We consider the task of discovering regulatory modules of genes from this gene expression data, but rather than simply learning a global clustering we will use the MCM to learn population specific clusterings of the genes. From around $20,000$ genes we filter down to the $N=1000$ most variable ones. The known tree over $T=7$ different populations is shown in Figure~\ref{fig:popTree}. Each data source $\mathbf{Y}^\tau$ is $N\times D^\tau$, where $D^\tau \subset D,  ~ \tau=1, \cdots, T$. We use this tree structure to define the covariance matrix and infer the branch length hyperparameters as described in Section~\ref{sec:kernels}. 

We again assess predictive performance on 10 heldout sets consisting of 10\% of data entries. In this case we find that although MCM performs better than independent clustering in each population, using a shared clustering of genes across all populations performs best (Table \ref{tab:results}). This suggests the biological conclusion that gene regulatory modules do not vary between diverse human populations. 

\section{Network modelling results}
\label{results}

We experimentally evaluate the ECS model on both synthetic and real-world data.

\subsection{Synthetic data}

\label{synthetic}

\begin{figure}[h]
\centering
\subfigure[Synthetic network data]{ \includegraphics[width=0.85\columnwidth]{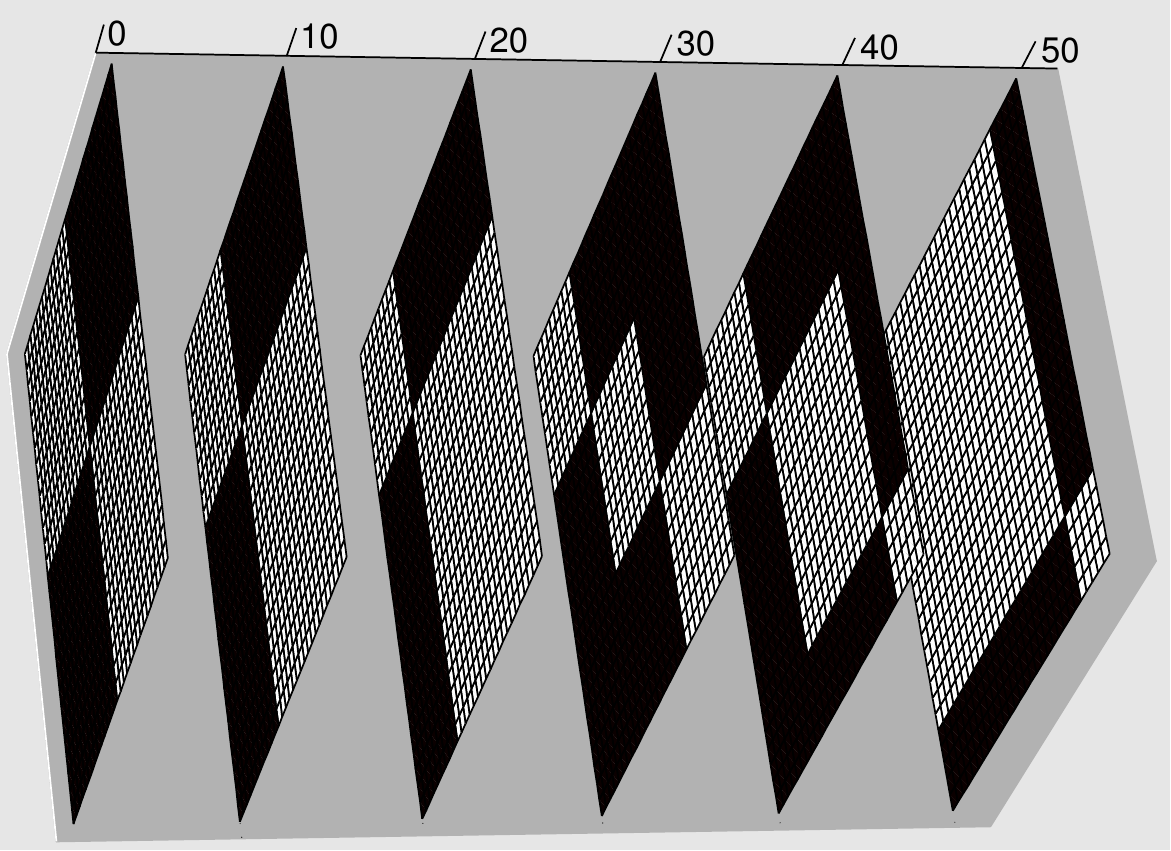}}
\subfigure[ECS partitioning]{\includegraphics[width=0.85\columnwidth]{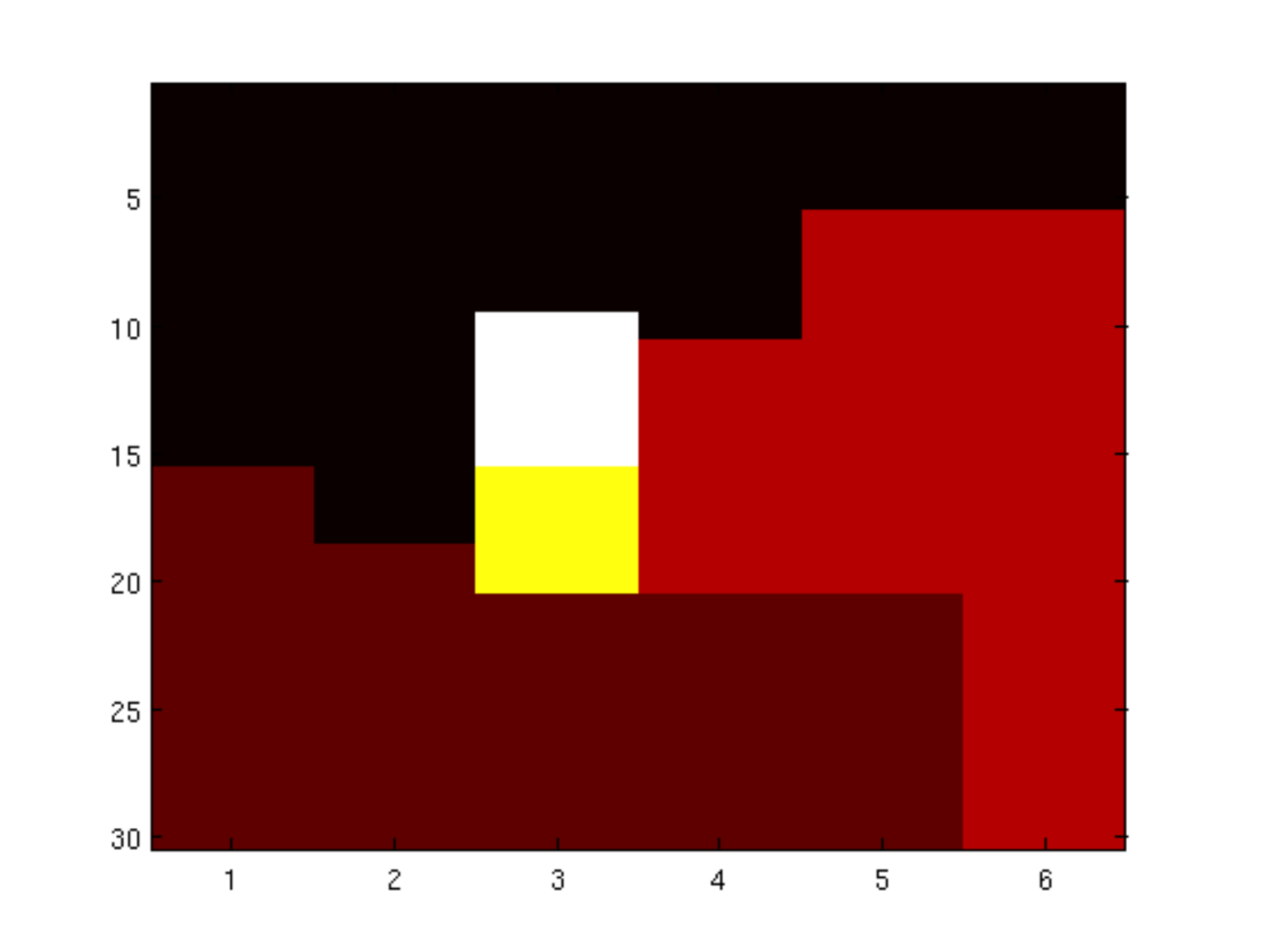}}
\vspace{-0.1in}
\caption{(a). Each matrix represents \textit{(non-)}links among pair of objects. White corresponds to one (link) and black to zero (\textit{non}-link) (b). Learnt clustering for synthetic network data. Colours denote assignment. 
\label{fig:timeVarfig}}
\end{figure}

We explore the ability of ECS to recover the partitions in synthetic time series network data. We hand-constructed a set of six square binary matrices which encode the friendship links among $N=30$ people evolving through time, as shown in Figure \ref{fig:timeVarfig}(a). People form groups which determine the links and non-links between them. As time passes, the partitioning of people changes; new friendship links are created while others break. The closer in time two snapshots are, the more similar we expect the related partitions will be. We ran ECS for 200 MCMC iterations and the sample with highest marginal likelihood is shown in Figure \ref{fig:timeVarfig}(b). The time-varying partition found roughly approximates the true group-stucture. We see that the solution provided by our model proposes two clusters at $t=0$, which shrink as a third cluster is generated between them. The partition found at $t=3$ is suboptimal: with more iterations, the white and yellow clusters used here may perhaps be 
replaced by the red cluster used at latter time points. However, multiple hypotheses are of course capable of explaining the same data. 

 

\subsection{van de Bunt's students}

In \citet{van1999friendship} 32 university freshman were surveyed at seven time points about who in their class they considered as friends. The first four time points were two weeks apart and the last four were three weeks apart. We binarise the original $0-5$ scale, taking `Best friendship', `Friendship' and `Friendly relationship' as $1$, `Neutral relationship', `Unknown person' and `Troubled relationship' as $0$, and `non-response' as missing. We also symmetrise the matrix by assuming friendship if either individual reported it. We test the predictive performance of ECS using the squared exponential kernel on this dataset by holding out 10\% of all links across all time points in 10 different training/test splits. We compare to independent IRMs at each time point and a model with shared clustering but independent link probabilities across all time points. The results in Table~\ref{tab:results} show that ECS significantly outperforms both these extremes in terms of heldout predictive performance. The 
average lengthscale learnt was $1.43$ weeks, showing that while there was similarity in community structure between proximal time points there were also significant changes to be taken into account over the time course of the full dataset. 

\section{Conclusion}
\label{conclusion}

Given the central role of clustering in unsupervised learning we expect the dependent partitioned-valued process we introduce here to have many potential applications. We have investigated two models derived from the DPVP: a multitask clustering model, which is, to the best of our knowledge, the first such model derived under a fully probabilistic framework, and a time series network model that is appropriate for the many real world networks that constantly evolve through time. 

Various directions for future work are open. Firstly, improved inference is of interest. While we used MCMC inference, variational methods would be a natural fit for either model: expectation propagation~\citep{Minka2001a} is known to be particularly effective for Gaussian process classification~\citep{Nickisch2008}, which is a subcomponent of DPVP, and variational Bayes~\citep{Ghahramani01propagationalgorithms} is commonly used for mixture modelling. 
Secondly, it is possible to have covariates associated with each object $n$. Making the $f$ also a function of these covariates would give a model related to the distance dependent CRP~\citep{blei2011distance}, and smart computation of the Cholesky would be only $O(T^3+N^3)$ rather than the naive $O(T^3N^3)$. Thirdly, other applications suggest themselves: modelling spatially varying ecological networks or the difference between regulatory modules across different human tissue types. Finally it would be of considerable interest to derive a dependent partition-valued process that, like the Fragmentation-Coagulation process, does not explicitly label clusters.


\setlength{\bibsep}{3pt}

{
\bibliography{dpm}
 }
\bibliographystyle{icml2013}

\end{document}